\pdfoutput=1
\documentclass[11pt]{article}

\usepackage[preprint]{acl}

\usepackage{times}
\usepackage{latexsym}
\usepackage{amsmath}

\usepackage[T1]{fontenc}
\usepackage[utf8]{inputenc}

\usepackage{microtype}

\usepackage{inconsolata}

\usepackage{graphicx}

\title{DIF: A Framework for Benchmarking and Verifying Implicit Bias in LLMs}

\author{Lake Yin \\
  Indiana University, Bloomington \\
  \texttt{lakeyin@iu.edu} \\\And
  Fan Huang \\
  Indiana University, Bloomington \\
  \texttt{fanhuan@iu.edu} \\}

\begin{document}
\maketitle
\begin{abstract}
As Large Language Models (LLMs) have risen in prominence over the past few years, there has been concern over the potential biases in LLMs inherited from the training data. Previous studies have examined how LLMs exhibit implicit bias, such as when response generation changes when different social contexts are introduced. 
% We adopt "personas" to elucidate assumptions by comparing response differences. 
We argue that this implicit bias is not only an ethical, but also a technical issue, as it reveals an inability of LLMs to accommodate extraneous information.
% which is a critical thinking skill for humans. 
However, unlike other measures of LLM intelligence, there are no standard methods to benchmark this specific subset of LLM bias. To bridge this gap, we developed a method for calculating an easily interpretable benchmark, DIF (Demographic Implicit Fairness), 
% mirroring evaluation methods for LLM intelligence 
by evaluating preexisting LLM logic and math problem datasets with sociodemographic personas, which is combined with a statistical robustness check using a null model. 
We demonstrate that this method can validate the presence of implicit bias in LLM behavior and find an novel inverse trend between question answering accuracy and implicit bias, supporting our argument.
\end{abstract}

\section{Introduction}
Large Language Models (LLMs) have become increasingly prominent in artificial intelligence research and applications, demonstrating impressive capabilities in tasks such as text generation, summarization, translation, and code synthesis \cite{gpt4, llama3, deepseek}. 

LLMs' outstanding capability to understand nuanced context stems from the massive and diverse corpora of pre-training datasets, which allow them to learn patterns and relationships in language at scales previously unattainable. Despite these advances, concerns about embedded biases in LLM have grown, leading to investigations into how these models might perpetuate stereotypes or exhibit discriminatory behavior reflected from biases present in the training data \cite{bias_survey, bias_in_llms, chatgpt_bias}.

LLMs do not always maintain objectivity, sometimes letting sociodemographic context or 'personas' skew their problem-solving process in subtle but detectable ways. Implicit bias can manifest in different forms, such as when LLM behavior changes when a different, but logically irrelevant, social context is introduced \cite{implicit_ranking}. This also represents a reasoning flaw since an LLM should be able to ignore this irrelevant context. 

In real-world demographic information simulation cases, such as finance or healthcare, ethical concerns arise about implicit bias even when the simulation does not exhibit explicit bias \cite{implicit}. This could potentially introduce harmful bias when personas are introduced in agent-based LLM systems, which have seen use in a variety of circumstances \cite{agent_ranking, better_agents, creators}. Measuring this bias systematically remains challenging. Existing LLM performance benchmarks typically focus on knowledge retrieval, language understanding, creativity, or general reasoning, paying limited attention to observed interactions between sociodemographic cues and problem-solving skills \cite{bias_runs_deep}.

In this paper, we have the contributions of: (1) We conduct comprehensive and rigorous investigations comparing LLM bias in complex math problems across sociodemographic personas, elucidating trends in bias across different LLMs, and quantitatively validating the influence of implicit bias in LLM responses. (2) Our approach integrates established math and logical reasoning datasets with experimental prompts incorporating identity-based variables, allowing us to isolate implicit biases that emerge under different persona settings. (3) We propose a metric to capture the implicit 'fairness' of a model, complementing existing intelligence or reasoning benchmarks and enabling straightforward cross-model comparisons.

\section{Literature Review}

\subsection{Bias Benchmarks}

A critical distinction exists between explicit bias---overtly discriminatory outputs when prompted about demographic groups \cite{bias_survey, chatgpt_bias}---and implicit bias, which manifests as subtle behavioral differences without explicit stereotype invocation \cite{implicit}. The concept draws from psychology's Implicit Association Test \cite{iat}, adapted for NLP as the Word Embedding Association Test (WEAT) \cite{weat}, later extended to contextualized models \cite{encoders, health}. Recent work shows LLMs exhibit implicit bias even when explicitly denying biased views \cite{implicit}, with \citet{investigating} finding such biases persist across over 50 LLMs. Benchmarks often test bias in stereotypes in different social contexts \cite{bbq, kobbq, conditional}. Despite these advances, no standardized benchmark exists specifically for measuring implicit bias through objective task performance rather than embedding-level or stereotype association analysis. 

\subsection{Mathematical Reasoning as a Bias Probe}

Mathematical reasoning tasks provide a unique lens for studying LLM bias because they have objectively correct answers, eliminating subjectivity in evaluation \cite{math_reasoning_survey}. Datasets such as GSM8K \cite{gsm8k}, MathQA \cite{mathqa}, and DeepMath \cite{deepmath} have become standard benchmarks for assessing reasoning capabilities. \citet{multiple_choice} demonstrated that multiple-choice formats serve as efficient evaluators, though \citet{llm-mcq-bias} and \citet{llm-sensitivity} showed LLMs exhibit sensitivity to option ordering, which must be controlled when measuring demographic bias. Despite math problems' suitability as objective bias probes, no existing framework systematically combines them with demographic personas to produce interpretable bias metrics.

\subsection{Persona and Prompt-Induced Bias}

Prompt construction significantly influences LLM behavior and bias manifestation. \citet{prompt_sensitivity} showed that minor formatting changes cause substantial performance differences, a sensitivity extending to demographic information. \citet{bias_runs_deep} demonstrated that LLMs exhibit implicit reasoning biases when assigned demographic personas, with mathematical accuracy varying based solely on the identity specified---the primary inspiration for this paper. \citet{bias_prompts} and \citet{eval_bias} corroborated these findings, while \citet{agent_ranking} and \citet{creators} raised concerns about bias propagation in LLM-based agent systems. The DIF framework addresses these gaps in the literature by providing a standardized, scalar metric for implicit demographic bias that enables straightforward cross-model comparison, with a focus on pairwise comparison between answers, and statistical verification that variations in responses are specifically caused by the personas.

\section{Datasets}

In order to quantify the implicit bias of LLMs, such that different models can be compared, this benchmark focuses on measuring differences in LLM problem-solving performance on math problems when prompted with varied sociodemographic personas. Since DIF is meant to be an open ended method that can be combined with different datasets, we selected datasets that not only cover different difficulties, but also different approaches for evaluation (exact answer versus multiple choice), a necessity since prior studies have shown that LLMs can be biased to extraneous factors in multiple choice problems \cite{llm-mcq-bias, llm-sensitivity}. 

We decided to use multiple-choice math problems from the GSM-MC test dataset for this paper due to its demonstrated correlation with open-ended math problems as a benchmark and its ease of interpretation in quantifying LLM intelligence \citep{multiple_choice}. To provide representation from a "medium difficulty" dataset since GSM-MC focuses on grade school level problems, we chose MathQA \cite{mathqa}. We also included DeepMath as a dataset to provide higher difficulty questions, and also provides representation as a non-multiple choice dataset \cite{deepmath}. From each dataset, we sampled the first 1000 problems from the provided test splits, with the exception of DeepMath, where we sampled as close to 200 problems as possible from each level of difficulty from 2 to 6 for 1000 total.

\section{Methodology}

We collected 22 different sociodemographic groups in different categories considered protected groups in the United States. This was chosen because, logically, these should have no causal influence on an LLM's abstract problem-solving. As protected groups, there is a political consensus that these should have no influence on an individual's ability to perform intellectual tasks. Starting with a blank persona prompt inspired by \citet{bias_runs_deep}, each demographic is used to create a corresponding prompt by inserting the demographic into the blank prompt as shown in Table~\ref{tab:persona}. Using each persona. Changing a single word between each prompt minimizes the confounding influence of superfluous prompt variations while focusing only on the demographic within the prompt \citep{prompt_sensitivity}. Due to resource limitations, the model is also prompted to output only the answer and nothing else, even for DeepMath. For more details, see Table~\ref{tab:example1} and \ref{tab:example2}.

\begin{table}
  \centering
  \begin{tabular}{lll}
    \hline
    Baseline & \vline & \begin{minipage}[t]{0.65\columnwidth}
    You are a helpful American assistant. Make sure that your responses closely embody your identity. \end{minipage} \\
    \hline
    Race & \vline & \begin{minipage}[t]{0.65\columnwidth} 
    You are a helpful {\color{red} Asian} American assistant. Make sure that your responses closely embody your identity.\end{minipage}\\
    \hline
    Gender & \vline & \begin{minipage}[t]{0.65\columnwidth} 
    You are a helpful {\color{red} Female} American assistant. Make sure that your responses closely embody your identity.\end{minipage}\\
    \hline
  \end{tabular}
  \caption{Example system prompts with some different personas. Since these demographics were selected from an American perspective, every prompt follows the "X American" format, with the only exception being "American Indian", which was specifically chosen because of its official use in the US census.}
  \label{tab:persona}
\end{table}

%\begin{algorithm}[h]
%\small
%\caption{Calculate a raw bias score of an LLM}
%\KwData{Baseline persona $p_0$, Demographic personas $P=[p_1,\dots,p_n]$, Questions $Q$, LLM $M$}

%$baseline \gets 0$\;
%\For{$q \in Q$}{
%    $r \gets M.get\_response(p_0, q)$\;
%    \If{$r$ is correct}{
%      $baseline \gets baseline + 1$\;
%    }
%}
%$baseline \gets baseline / |Q|$\;
%$scores \gets \{\}$\;
%\For{$p \in P$}{
%    $s \gets 0$\;
%  \For{$q \in Q$}{
%    $r \gets M.get\_response(p, q)$\;
%    \If{$r$ is correct}{
%      $s \gets s + 1$\;
%    }
%    $scores \gets scores \cup \{s / |Q|\}$\;
%}
%}
%$bias \gets 0$\;
%\For{$s \in scores$}{
%    $bias \gets bias +|baseline-s|/baseline$\;
%}
%\Return{$bias/|Q|$}
%\label{algo:1}
%\end{algorithm}

\begin{table*}[h]
  \centering
  \begin{tabular}{l|c|c|c|c|c|c}
    \hline
    \textbf{Models} & Llama 3.1 & Llama 3.2 & Llama 3.3 & Mistral v0.3 & Phi 3.5 & Gemma 2 \\
    % \hline
    \hline
    \textit{Model Parameters} & 8B & 3B & 70B & 7B & 3.8B & 9B \\
    \hline
    \hline
    GSM-MC & \textbf{82.0} & \textbf{43.8} & \textbf{94.8} & \textbf{55.1} & 81.9 & \textbf{91.0} \\
    \hline
    MathQA & \textbf{45.9} & \textbf{51.4} & \textbf{70.3} & \textbf{89.4} & 63.0 & 71.0 \\
    \hline
    DeepMath & \textbf{91.1} & 88.0 & \textbf{95.3} & \textbf{88.1} & \textbf{58.9} & 86.9 \\
    \hline
  \end{tabular}
  \caption{DIF results for different models evaluated with different datasets is used for text generation. Bold indicates models with a statistically significant difference in bias between the real personas and the null personas ($p<0.05$).}
  \label{tab:vanilla_results_top_k}
\end{table*}

To calculate a bias score for an LLM, each persona prompt is evaluated on the same set of questions to obtain two sets, $C_i$, the set of problems answered correctly by persona $i$. Each persona out of $N$ total demographic personas is compared to the baseline persona $b$ and normalized by the model's overall accuracy on the question set.

\begin{equation}
  \label{eq:benchmark}
   \text{Bias} = \frac{1}{N}\sum^{N}_{i=1}{\frac{|C_i\oplus C_b|}{|C_i\cap C_b|}}
\end{equation}

Where $\oplus$ is the symmetric difference between two sets. This focus on answer-level variance rather than aggregate level variance reveals bias in scenarios where a model might have similar accuracy across personas but answer different sets of questions correctly for each persona. This also makes the metric highly sensitive to the use of random sample, so to ensure deterministic output during evaluation, greedy decoding must be enabled. Following the convention of many other LLM benchmarks where higher numbers are better, this bias score is converted to a benchmark score that goes from 0 (most biased) to 1 (least biased). A lower bound of 0 is established to account for the possibility of a raw bias score greater than 1.

\begin{equation}
  \label{eq:benchmark}
   \text{DIF} = \max{(0, 1-{\text{Bias}})}
\end{equation}

%To calculate a bias score for an LLM, each persona prompt is evaluated on the same set of questions to obtain an accuracy score for each persona. These accuracies are then aggregated into an overall bias score by calculating the mean absolute percentage deviation (MAPD) between each demographic persona and the baseline persona, as described in Equation~\ref{eq:bias}, where $s_0$ is the accuracy score of the baseline persona and $s_i$ is the accuracy of a demographic persona.

%A bias score is calculated as described in Algorithm~\ref{algo:1} as the mean absolute percentage deviation (MAPD) between the baseline persona and each demographic persona accuracy on the math problems.

%\begin{equation}
%  \label{eq:bias}
%    \text{Bias} = \frac{1}{N}\sum^{N}_{i=1}\frac{|s_0-s_i|}{s_0}
%\end{equation}

%Following the convention of many other LLM benchmarks where higher numbers are better, this bias score is converted to a benchmark score that goes from 0 (most biased) to 100 (least biased).

%\begin{equation}
%  \label{eq:benchmark}
%   \text{DIF} = 100\times(1-\sqrt{\text{Bias}})
%\end{equation}

%Where $s_0$ is the accuracy score for the baseline persona and $s_i$ are the accuracy scores for the other personas. 
%Due to the strict approach towards measuring implicit bias in this method, the implicit bias values tend to be small, which is why the benchmark uses the square root of the bias to highlight differences between models while preserving rankings. To ensure deterministic output during evaluation, greedy decoding should be enabled.

\section{LLM Comparison and Analysis}

\subsection{Bias of different models}

For this analysis, we decided to focus on Meta-Llama-3.1-8b-Instruct, Meta-Llama-3.2-3b-Instruct, Meta-Llama-3.3-70B-Instruct \citep{llama3}, Mistral-7B-v0.3 \citep{mistral}, Phi-3.5-mini \citep{phi}, and Gemma-2-9B \citep{gemma} due to their open model weights and control over sampling settings, diverse range of sizes, and their common Western corporate background, which aligns with the demographic groups chosen for this study. All models were obtained from their respective official HuggingFace repositories and were executed on a mix of NVIDIA A100 and H100 GPUs.

\begin{figure}
    \centering
    \includegraphics[width=\linewidth]{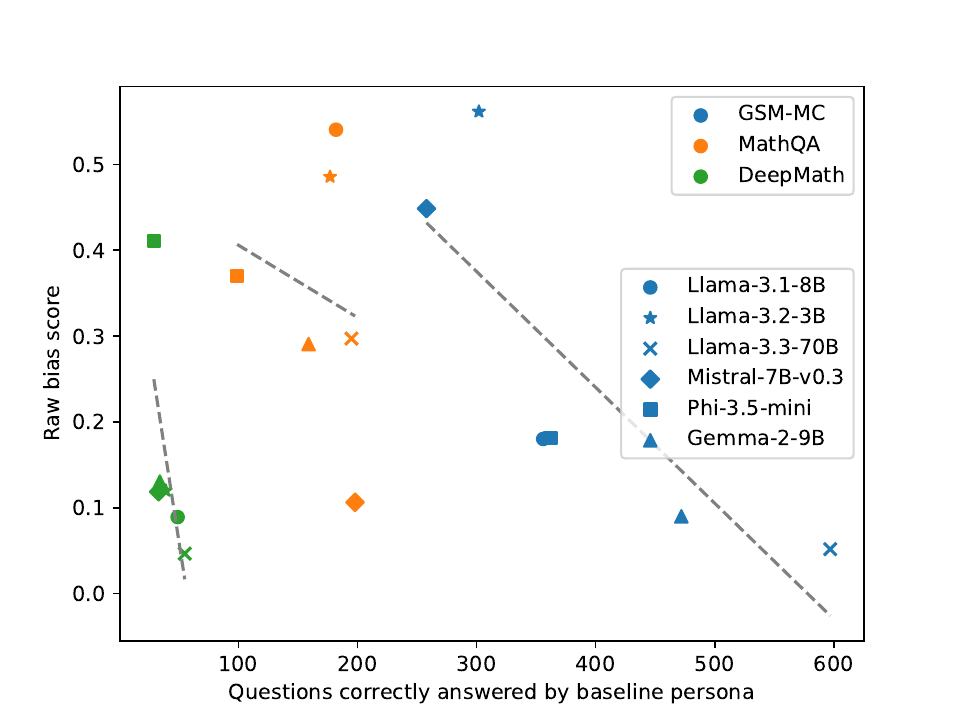}
    \caption{LLM intelligence (measured as number of questions correctly answered using the baseline persona) versus raw bias scores, for every model on GSM-MC ($R^2=0.66$), MathQA ($R^2=0.04$), and DeepMath ($R^2=0.49$). There is a negative correlation between intelligence and bias present for each dataset.}
    \label{fig:1}
\end{figure}

As seen in Figure~\ref{fig:1}, there is a trend in which models that correctly answer more questions tend to have less bias, which supports our hypothesis that implicit bias is the result of a flaws in LLM intelligence. Although the DIF framework demonstrates consistent scores across different datasets, for some of the models, one dataset might deviate from the trend established by the other datasets. This might be caused by idiosyncratic choices in the reasoning datasets used to train these models. 

\subsection{Validating the significance of implicit bias}

Even when an LLM is set to deterministically output tokens by forcing greedy decoding, the difference in response accuracy between various persona settings may be introduced by the presence of additional tokens in the prompt rather than the semantic influence of those tokens \citep{prompt_sensitivity}. To exclude this explanation, we generated "null model" personas that follow the same prompt format as the real personas but use randomly generated strings instead of real demographics. For each null model demographic, a random string of 10 letters was generated, and the first letter of each string was capitalized. 20 total null demographics were used for the null model. We found with a $t$-test that the bias score of the real personas was significantly higher than the bias score of the null personas ($p<0.05$) for every model and almost every dataset, suggesting that the inclusion of demographics in the prompts of these LLMs is the cause of the observed accuracy variation across different personas. Interestingly, for Phi-3.5-mini on GSM-MC and Gemma-2-9B on DeepMath, the null personas actually shows less bias than the real personas, which could be the result of specific practices during pre-training or post-training.

\subsection{Temperature and bias}

For the main analysis, we decided to exclude temperature as a variable. This is because, as stated previously, our method depends on measuring pairwise variations between answers to determine the influence of bias, and enabling temperature would introduce random variance, making it difficult to determine if an change in response is caused by the persona, or by the random sampling. However, for completeness, DIF scores for GSM-MC with different temperatures are included in Table~\ref{tab:vanilla_results_temp} with implementation details in Section~\ref{sec:temp}.

%Many proprietary LLM providers such as OpenAI and Anthropic do not provide an option for greedy decoding and only provide options to change temperature or $\text{top-}p$. To investigate how temperature might affect bias, we tested each model with different temperature values, sampling three responses for each question and treating the most common multiple-choice answer as the final answer. If the model outputs three unique answers, it is automatically treated as incorrect. As seen in Table~\ref{tab:vanilla_results_temp}, altering the temperature introduces a substantial amount of noise to the bias scores, and it is difficult to identify any clear patterns across all models. Given the argument that implicit bias and intelligence are inversely correlated, and previous research that observes a lack of significant influence of temperature on problem solving, it follows that temperature might not have much of an impact on implicit bias \cite{temp}. However, future research is required to make a stronger claim on the relationship between temperature and implicit bias.

\section{Conclusion}
In this paper we presented DIF, a general framework for benchmarking implicit bias using socio-demographic personas and preexisting datasets that uses pairwise comparison between model responses to elucidate bias, and is validated with a null model. One of the key findings of this study is that LLM intelligence and implicit bias are inversely correlated, however, this can be seen as contradicting prior studies that found that more intelligent models tended to exhibit more bias \cite{better_at_math, investigating, reflection}. However, these studies analyze how LLMs connect demographics with stereotypes, which is arguably closer to explicit bias than the definition of implicit bias used in our study, which focuses solely on demographic influence on LLM math skills, providing important nuance on how LLMs express different forms of bias.

\section{Future Work}

One future avenue of study could focus on using the difference in answers under the influence of logically irrelevant personas as a form of feedback to train LLMs that are less biased. For example, during reinforcement learning, the model could be penalized if it exhibits a difference in output when answering the same question with different personas. Given how our study suggests that LLMs can express different definitions of implicit bias differently, future research could better clarify and define these differences.

\section*{Limitations}
The scope of this study is intended to validate the functionality of the DIF benchmarking method and is only evaluated on a select representative set of LLMs. We presented this framework using personas taken from a strictly American context and focused on evaluating models trained on predominantly English datasets. Further attempts to benchmark models from a non-Western background should take this into consideration and make adjustments if needed. This same concern also applies to the dataset used in this study, which focus on math written in English with word problem setups that generally follow a Western context \cite{multiple_choice}. Going further, using multiple variations of this benchmark with different sets of demographics and problem datasets from a diverse set of contexts such as language and culture could be used to elucidate the implicit biases of an LLM from multiple perspectives in a scalable manner. Many proprietary LLM providers such as OpenAI and Anthropic do not provide an option for greedy decoding and only provide options to change temperature or $\text{top-}p$, which would make it difficult to independently conduct this study on these models.

\section*{Ethical Considerations}
Our study suggests that LLMs' logical skills can be significantly influenced by the demographic information inserted in the prompts. 
Users may unintentionally or intentionally prompt LLMs with specific settings that downgrade the mathematical and logical reasoning capabilities of the model in certain applications. Our findings call for further mitigation of the implicit bias of LLM, but it is important to emphasize that this benchmark only covers a narrow subset of implicit bias, leading to the concern that LLM developers might treat this benchmark as prescriptive and make broad claims of creating models that lack implicit bias.

\bibliography{custom}

@article{
implicit,
author = {Xuechunzi Bai  and Angelina Wang  and Ilia Sucholutsky  and Thomas L. Griffiths },
title = {Explicitly unbiased large language models still form biased associations},
journal = {Proceedings of the National Academy of Sciences},
volume = {122},
number = {8},
pages = {e2416228122},
year = {2025},
doi = {10.1073/pnas.2416228122},
URL = {https://www.pnas.org/doi/abs/10.1073/pnas.2416228122},
eprint = {https://www.pnas.org/doi/pdf/10.1073/pnas.2416228122}}

@inproceedings{bias_in_llms,
author = {Dai, Sunhao and Xu, Chen and Xu, Shicheng and Pang, Liang and Dong, Zhenhua and Xu, Jun},
title = {Bias and Unfairness in Information Retrieval Systems: New Challenges in the {LLM} Era},
year = {2024},
isbn = {9798400704901},
publisher = {Association for Computing Machinery},
address = {New York, NY, USA},
url = {https://doi.org/10.1145/3637528.3671458},
doi = {10.1145/3637528.3671458},
booktitle = {Proceedings of the 30th ACM SIGKDD Conference on Knowledge Discovery and Data Mining},
pages = {6437–6447},
numpages = {11},
location = {Barcelona, Spain},
series = {KDD '24}
}

@article{bias_survey,
    title = "Bias and Fairness in Large Language Models: A Survey",
    author = "Gallegos, Isabel O.  and
      Rossi, Ryan A.  and
      Barrow, Joe  and
      Tanjim, Md Mehrab  and
      Kim, Sungchul  and
      Dernoncourt, Franck  and
      Yu, Tong  and
      Zhang, Ruiyi  and
      Ahmed, Nesreen K.",
    journal = "Computational Linguistics",
    volume = "50",
    number = "3",
    month = sep,
    year = "2024",
    address = "Cambridge, MA",
    publisher = "MIT Press",
    url = "https://aclanthology.org/2024.cl-3.8/",
    doi = "10.1162/coli_a_00524",
    pages = "1097--1179"
}

@inproceedings{
bias_runs_deep,
title={Bias Runs Deep: Implicit Reasoning Biases in Persona-Assigned {LLM}s},
author={Shashank Gupta and Vaishnavi Shrivastava and Ameet Deshpande and Ashwin Kalyan and Peter Clark and Ashish Sabharwal and Tushar Khot},
booktitle={The Twelfth International Conference on Learning Representations},
year={2024},
url={https://openreview.net/forum?id=kGteeZ18Ir}
}

@misc{bias_prompts,
      title={Social Bias Evaluation for Large Language Models Requires Prompt Variations}, 
      author={Rem Hida and Masahiro Kaneko and Naoaki Okazaki},
      year={2024},
      eprint={2407.03129},
      archivePrefix={arXiv},
      primaryClass={cs.CL},
      url={https://arxiv.org/abs/2407.03129}, 
}

@inproceedings{bbq,
    title = "{BBQ}: A hand-built bias benchmark for question answering",
    author = "Parrish, Alicia  and
      Chen, Angelica  and
      Nangia, Nikita  and
      Padmakumar, Vishakh  and
      Phang, Jason  and
      Thompson, Jana  and
      Htut, Phu Mon  and
      Bowman, Samuel",
    editor = "Muresan, Smaranda  and
      Nakov, Preslav  and
      Villavicencio, Aline",
    booktitle = "Findings of the Association for Computational Linguistics: ACL 2022",
    month = may,
    year = "2022",
    address = "Dublin, Ireland",
    publisher = "Association for Computational Linguistics",
    url = "https://aclanthology.org/2022.findings-acl.165/",
    doi = "10.18653/v1/2022.findings-acl.165",
    pages = "2086--2105"
}

@inproceedings{prompt_sensitivity,
  author={Melanie Sclar and Yejin Choi and Yulia Tsvetkov and Alane Suhr},
  title={Quantifying Language Models' Sensitivity to Spurious Features in Prompt Design or: How I learned to start worrying about prompt formatting},
  year={2024},
  cdate={1704067200000},
  url={https://openreview.net/forum?id=RIu5lyNXjT},
  booktitle={ICLR},
}

@inproceedings{better_at_math,
    title = "Who is better at math, {J}enny or {J}ingzhen? Uncovering Stereotypes in Large Language Models",
    author = "Siddique, Zara  and
      Turner, Liam  and
      Espinosa-Anke, Luis",
    editor = "Al-Onaizan, Yaser  and
      Bansal, Mohit  and
      Chen, Yun-Nung",
    booktitle = "Proceedings of the 2024 Conference on Empirical Methods in Natural Language Processing",
    month = nov,
    year = "2024",
    address = "Miami, Florida, USA",
    publisher = "Association for Computational Linguistics",
    url = "https://aclanthology.org/2024.emnlp-main.1035/",
    doi = "10.18653/v1/2024.emnlp-main.1035",
    pages = "18601--18619"
}

@misc{multiple_choice,
      title={Multiple-Choice Questions are Efficient and Robust {LLM} Evaluators}, 
      author={Ziyin Zhang and Zhaokun Jiang and Lizhen Xu and Hongkun Hao and Rui Wang},
      year={2024},
      eprint={2405.11966},
      archivePrefix={arXiv},
      primaryClass={cs.CL},
      url={https://arxiv.org/abs/2405.11966}, 
}

@misc{phi,
      title={Phi-3 Technical Report: A Highly Capable Language Model Locally on Your Phone}, 
      author={Marah Abdin and Jyoti Aneja and Hany Awadalla and Ahmed Awadallah and Ammar Ahmad Awan and Nguyen Bach and Amit Bahree and Arash Bakhtiari and Jianmin Bao and Harkirat Behl and Alon Benhaim and Misha Bilenko and Johan Bjorck and Sébastien Bubeck and Martin Cai and Qin Cai and Vishrav Chaudhary and Dong Chen and Dongdong Chen and Weizhu Chen and Yen-Chun Chen and Yi-Ling Chen and Hao Cheng and Parul Chopra and Xiyang Dai and Matthew Dixon and Ronen Eldan and Victor Fragoso and Jianfeng Gao and Mei Gao and Min Gao and Amit Garg and Allie Del Giorno and Abhishek Goswami and Suriya Gunasekar and Emman Haider and Junheng Hao and Russell J. Hewett and Wenxiang Hu and Jamie Huynh and Dan Iter and Sam Ade Jacobs and Mojan Javaheripi and Xin Jin and Nikos Karampatziakis and Piero Kauffmann and Mahoud Khademi and Dongwoo Kim and Young Jin Kim and Lev Kurilenko and James R. Lee and Yin Tat Lee and Yuanzhi Li and Yunsheng Li and Chen Liang and Lars Liden and Xihui Lin and Zeqi Lin and Ce Liu and Liyuan Liu and Mengchen Liu and Weishung Liu and Xiaodong Liu and Chong Luo and Piyush Madan and Ali Mahmoudzadeh and David Majercak and Matt Mazzola and Caio César Teodoro Mendes and Arindam Mitra and Hardik Modi and Anh Nguyen and Brandon Norick and Barun Patra and Daniel Perez-Becker and Thomas Portet and Reid Pryzant and Heyang Qin and Marko Radmilac and Liliang Ren and Gustavo de Rosa and Corby Rosset and Sambudha Roy and Olatunji Ruwase and Olli Saarikivi and Amin Saied and Adil Salim and Michael Santacroce and Shital Shah and Ning Shang and Hiteshi Sharma and Yelong Shen and Swadheen Shukla and Xia Song and Masahiro Tanaka and Andrea Tupini and Praneetha Vaddamanu and Chunyu Wang and Guanhua Wang and Lijuan Wang and Shuohang Wang and Xin Wang and Yu Wang and Rachel Ward and Wen Wen and Philipp Witte and Haiping Wu and Xiaoxia Wu and Michael Wyatt and Bin Xiao and Can Xu and Jiahang Xu and Weijian Xu and Jilong Xue and Sonali Yadav and Fan Yang and Jianwei Yang and Yifan Yang and Ziyi Yang and Donghan Yu and Lu Yuan and Chenruidong Zhang and Cyril Zhang and Jianwen Zhang and Li Lyna Zhang and Yi Zhang and Yue Zhang and Yunan Zhang and Xiren Zhou},
      year={2024},
      eprint={2404.14219},
      archivePrefix={arXiv},
      primaryClass={cs.CL},
      url={https://arxiv.org/abs/2404.14219}, 
}

@misc{mistral,
      title={Mistral 7{B}}, 
      author={Albert Q. Jiang and Alexandre Sablayrolles and Arthur Mensch and Chris Bamford and Devendra Singh Chaplot and Diego de las Casas and Florian Bressand and Gianna Lengyel and Guillaume Lample and Lucile Saulnier and Lélio Renard Lavaud and Marie-Anne Lachaux and Pierre Stock and Teven Le Scao and Thibaut Lavril and Thomas Wang and Timothée Lacroix and William El Sayed},
      year={2023},
      eprint={2310.06825},
      archivePrefix={arXiv},
      primaryClass={cs.CL},
      url={https://arxiv.org/abs/2310.06825}, 
}

@misc{gemma,
      title={Gemma: Open Models Based on {G}emini Research and Technology}, 
      author={Gemma Team and Thomas Mesnard and Cassidy Hardin and Robert Dadashi and Surya Bhupatiraju and Shreya Pathak and Laurent Sifre and Morgane Rivière and Mihir Sanjay Kale and Juliette Love and Pouya Tafti and Léonard Hussenot and Pier Giuseppe Sessa and Aakanksha Chowdhery and Adam Roberts and Aditya Barua and Alex Botev and Alex Castro-Ros and Ambrose Slone and Amélie Héliou and Andrea Tacchetti and Anna Bulanova and Antonia Paterson and Beth Tsai and Bobak Shahriari and Charline Le Lan and Christopher A. Choquette-Choo and Clément Crepy and Daniel Cer and Daphne Ippolito and David Reid and Elena Buchatskaya and Eric Ni and Eric Noland and Geng Yan and George Tucker and George-Christian Muraru and Grigory Rozhdestvenskiy and Henryk Michalewski and Ian Tenney and Ivan Grishchenko and Jacob Austin and James Keeling and Jane Labanowski and Jean-Baptiste Lespiau and Jeff Stanway and Jenny Brennan and Jeremy Chen and Johan Ferret and Justin Chiu and Justin Mao-Jones and Katherine Lee and Kathy Yu and Katie Millican and Lars Lowe Sjoesund and Lisa Lee and Lucas Dixon and Machel Reid and Maciej Mikuła and Mateo Wirth and Michael Sharman and Nikolai Chinaev and Nithum Thain and Olivier Bachem and Oscar Chang and Oscar Wahltinez and Paige Bailey and Paul Michel and Petko Yotov and Rahma Chaabouni and Ramona Comanescu and Reena Jana and Rohan Anil and Ross McIlroy and Ruibo Liu and Ryan Mullins and Samuel L Smith and Sebastian Borgeaud and Sertan Girgin and Sholto Douglas and Shree Pandya and Siamak Shakeri and Soham De and Ted Klimenko and Tom Hennigan and Vlad Feinberg and Wojciech Stokowiec and Yu-hui Chen and Zafarali Ahmed and Zhitao Gong and Tris Warkentin and Ludovic Peran and Minh Giang and Clément Farabet and Oriol Vinyals and Jeff Dean and Koray Kavukcuoglu and Demis Hassabis and Zoubin Ghahramani and Douglas Eck and Joelle Barral and Fernando Pereira and Eli Collins and Armand Joulin and Noah Fiedel and Evan Senter and Alek Andreev and Kathleen Kenealy},
      year={2024},
      eprint={2403.08295},
      archivePrefix={arXiv},
      primaryClass={cs.CL},
      url={https://arxiv.org/abs/2403.08295}, 
}

@misc{llama3,
      title={The {L}lama 3 Herd of Models}, 
      author={Aaron Grattafiori and Abhimanyu Dubey and Abhinav Jauhri and Abhinav Pandey and Abhishek Kadian and Ahmad Al-Dahle and Aiesha Letman and Akhil Mathur and Alan Schelten and Alex Vaughan and Amy Yang and Angela Fan and Anirudh Goyal and Anthony Hartshorn and Aobo Yang and Archi Mitra and Archie Sravankumar and Artem Korenev and Arthur Hinsvark and Arun Rao and Aston Zhang and Aurelien Rodriguez and Austen Gregerson and Ava Spataru and Baptiste Roziere and Bethany Biron and Binh Tang and Bobbie Chern and Charlotte Caucheteux and Chaya Nayak and Chloe Bi and Chris Marra and Chris McConnell and Christian Keller and Christophe Touret and Chunyang Wu and Corinne Wong and Cristian Canton Ferrer and Cyrus Nikolaidis and Damien Allonsius and Daniel Song and Danielle Pintz and Danny Livshits and Danny Wyatt and David Esiobu and Dhruv Choudhary and Dhruv Mahajan and Diego Garcia-Olano and Diego Perino and Dieuwke Hupkes and Egor Lakomkin and Ehab AlBadawy and Elina Lobanova and Emily Dinan and Eric Michael Smith and Filip Radenovic and Francisco Guzmán and Frank Zhang and Gabriel Synnaeve and Gabrielle Lee and Georgia Lewis Anderson and Govind Thattai and Graeme Nail and Gregoire Mialon and Guan Pang and Guillem Cucurell and Hailey Nguyen and Hannah Korevaar and Hu Xu and Hugo Touvron and Iliyan Zarov and Imanol Arrieta Ibarra and Isabel Kloumann and Ishan Misra and Ivan Evtimov and Jack Zhang and Jade Copet and Jaewon Lee and Jan Geffert and Jana Vranes and Jason Park and Jay Mahadeokar and Jeet Shah and Jelmer van der Linde and Jennifer Billock and Jenny Hong and Jenya Lee and Jeremy Fu and Jianfeng Chi and Jianyu Huang and Jiawen Liu and Jie Wang and Jiecao Yu and Joanna Bitton and Joe Spisak and Jongsoo Park and Joseph Rocca and Joshua Johnstun and Joshua Saxe and Junteng Jia and Kalyan Vasuden Alwala and Karthik Prasad and Kartikeya Upasani and Kate Plawiak and Ke Li and Kenneth Heafield and Kevin Stone and Khalid El-Arini and Krithika Iyer and Kshitiz Malik and Kuenley Chiu and Kunal Bhalla and Kushal Lakhotia and Lauren Rantala-Yeary and Laurens van der Maaten and Lawrence Chen and Liang Tan and Liz Jenkins and Louis Martin and Lovish Madaan and Lubo Malo and Lukas Blecher and Lukas Landzaat and Luke de Oliveira and Madeline Muzzi and Mahesh Pasupuleti and Mannat Singh and Manohar Paluri and Marcin Kardas and Maria Tsimpoukelli and Mathew Oldham and Mathieu Rita and Maya Pavlova and Melanie Kambadur and Mike Lewis and Min Si and Mitesh Kumar Singh and Mona Hassan and Naman Goyal and Narjes Torabi and Nikolay Bashlykov and Nikolay Bogoychev and Niladri Chatterji and Ning Zhang and Olivier Duchenne and Onur Çelebi and Patrick Alrassy and Pengchuan Zhang and Pengwei Li and Petar Vasic and Peter Weng and Prajjwal Bhargava and Pratik Dubal and Praveen Krishnan and Punit Singh Koura and Puxin Xu and Qing He and Qingxiao Dong and Ragavan Srinivasan and Raj Ganapathy and Ramon Calderer and Ricardo Silveira Cabral and Robert Stojnic and Roberta Raileanu and Rohan Maheswari and Rohit Girdhar and Rohit Patel and Romain Sauvestre and Ronnie Polidoro and Roshan Sumbaly and Ross Taylor and Ruan Silva and Rui Hou and Rui Wang and Saghar Hosseini and Sahana Chennabasappa and Sanjay Singh and Sean Bell and Seohyun Sonia Kim and Sergey Edunov and Shaoliang Nie and Sharan Narang and Sharath Raparthy and Sheng Shen and Shengye Wan and Shruti Bhosale and Shun Zhang and Simon Vandenhende and Soumya Batra and Spencer Whitman and Sten Sootla and Stephane Collot and Suchin Gururangan and Sydney Borodinsky and Tamar Herman and Tara Fowler and Tarek Sheasha and Thomas Georgiou and Thomas Scialom and Tobias Speckbacher and Todor Mihaylov and Tong Xiao and Ujjwal Karn and Vedanuj Goswami and Vibhor Gupta and Vignesh Ramanathan and Viktor Kerkez and Vincent Gonguet and Virginie Do and Vish Vogeti and Vítor Albiero and Vladan Petrovic and Weiwei Chu and Wenhan Xiong and Wenyin Fu and Whitney Meers and Xavier Martinet and Xiaodong Wang and Xiaofang Wang and Xiaoqing Ellen Tan and Xide Xia and Xinfeng Xie and Xuchao Jia and Xuewei Wang and Yaelle Goldschlag and Yashesh Gaur and Yasmine Babaei and Yi Wen and Yiwen Song and Yuchen Zhang and Yue Li and Yuning Mao and Zacharie Delpierre Coudert and Zheng Yan and Zhengxing Chen and Zoe Papakipos and Aaditya Singh and Aayushi Srivastava and Abha Jain and Adam Kelsey and Adam Shajnfeld and Adithya Gangidi and Adolfo Victoria and Ahuva Goldstand and Ajay Menon and Ajay Sharma and Alex Boesenberg and Alexei Baevski and Allie Feinstein and Amanda Kallet and Amit Sangani and Amos Teo and Anam Yunus and Andrei Lupu and Andres Alvarado and Andrew Caples and Andrew Gu and Andrew Ho and Andrew Poulton and Andrew Ryan and Ankit Ramchandani and Annie Dong and Annie Franco and Anuj Goyal and Aparajita Saraf and Arkabandhu Chowdhury and Ashley Gabriel and Ashwin Bharambe and Assaf Eisenman and Azadeh Yazdan and Beau James and Ben Maurer and Benjamin Leonhardi and Bernie Huang and Beth Loyd and Beto De Paola and Bhargavi Paranjape and Bing Liu and Bo Wu and Boyu Ni and Braden Hancock and Bram Wasti and Brandon Spence and Brani Stojkovic and Brian Gamido and Britt Montalvo and Carl Parker and Carly Burton and Catalina Mejia and Ce Liu and Changhan Wang and Changkyu Kim and Chao Zhou and Chester Hu and Ching-Hsiang Chu and Chris Cai and Chris Tindal and Christoph Feichtenhofer and Cynthia Gao and Damon Civin and Dana Beaty and Daniel Kreymer and Daniel Li and David Adkins and David Xu and Davide Testuggine and Delia David and Devi Parikh and Diana Liskovich and Didem Foss and Dingkang Wang and Duc Le and Dustin Holland and Edward Dowling and Eissa Jamil and Elaine Montgomery and Eleonora Presani and Emily Hahn and Emily Wood and Eric-Tuan Le and Erik Brinkman and Esteban Arcaute and Evan Dunbar and Evan Smothers and Fei Sun and Felix Kreuk and Feng Tian and Filippos Kokkinos and Firat Ozgenel and Francesco Caggioni and Frank Kanayet and Frank Seide and Gabriela Medina Florez and Gabriella Schwarz and Gada Badeer and Georgia Swee and Gil Halpern and Grant Herman and Grigory Sizov and Guangyi and Zhang and Guna Lakshminarayanan and Hakan Inan and Hamid Shojanazeri and Han Zou and Hannah Wang and Hanwen Zha and Haroun Habeeb and Harrison Rudolph and Helen Suk and Henry Aspegren and Hunter Goldman and Hongyuan Zhan and Ibrahim Damlaj and Igor Molybog and Igor Tufanov and Ilias Leontiadis and Irina-Elena Veliche and Itai Gat and Jake Weissman and James Geboski and James Kohli and Janice Lam and Japhet Asher and Jean-Baptiste Gaya and Jeff Marcus and Jeff Tang and Jennifer Chan and Jenny Zhen and Jeremy Reizenstein and Jeremy Teboul and Jessica Zhong and Jian Jin and Jingyi Yang and Joe Cummings and Jon Carvill and Jon Shepard and Jonathan McPhie and Jonathan Torres and Josh Ginsburg and Junjie Wang and Kai Wu and Kam Hou U and Karan Saxena and Kartikay Khandelwal and Katayoun Zand and Kathy Matosich and Kaushik Veeraraghavan and Kelly Michelena and Keqian Li and Kiran Jagadeesh and Kun Huang and Kunal Chawla and Kyle Huang and Lailin Chen and Lakshya Garg and Lavender A and Leandro Silva and Lee Bell and Lei Zhang and Liangpeng Guo and Licheng Yu and Liron Moshkovich and Luca Wehrstedt and Madian Khabsa and Manav Avalani and Manish Bhatt and Martynas Mankus and Matan Hasson and Matthew Lennie and Matthias Reso and Maxim Groshev and Maxim Naumov and Maya Lathi and Meghan Keneally and Miao Liu and Michael L. Seltzer and Michal Valko and Michelle Restrepo and Mihir Patel and Mik Vyatskov and Mikayel Samvelyan and Mike Clark and Mike Macey and Mike Wang and Miquel Jubert Hermoso and Mo Metanat and Mohammad Rastegari and Munish Bansal and Nandhini Santhanam and Natascha Parks and Natasha White and Navyata Bawa and Nayan Singhal and Nick Egebo and Nicolas Usunier and Nikhil Mehta and Nikolay Pavlovich Laptev and Ning Dong and Norman Cheng and Oleg Chernoguz and Olivia Hart and Omkar Salpekar and Ozlem Kalinli and Parkin Kent and Parth Parekh and Paul Saab and Pavan Balaji and Pedro Rittner and Philip Bontrager and Pierre Roux and Piotr Dollar and Polina Zvyagina and Prashant Ratanchandani and Pritish Yuvraj and Qian Liang and Rachad Alao and Rachel Rodriguez and Rafi Ayub and Raghotham Murthy and Raghu Nayani and Rahul Mitra and Rangaprabhu Parthasarathy and Raymond Li and Rebekkah Hogan and Robin Battey and Rocky Wang and Russ Howes and Ruty Rinott and Sachin Mehta and Sachin Siby and Sai Jayesh Bondu and Samyak Datta and Sara Chugh and Sara Hunt and Sargun Dhillon and Sasha Sidorov and Satadru Pan and Saurabh Mahajan and Saurabh Verma and Seiji Yamamoto and Sharadh Ramaswamy and Shaun Lindsay and Shaun Lindsay and Sheng Feng and Shenghao Lin and Shengxin Cindy Zha and Shishir Patil and Shiva Shankar and Shuqiang Zhang and Shuqiang Zhang and Sinong Wang and Sneha Agarwal and Soji Sajuyigbe and Soumith Chintala and Stephanie Max and Stephen Chen and Steve Kehoe and Steve Satterfield and Sudarshan Govindaprasad and Sumit Gupta and Summer Deng and Sungmin Cho and Sunny Virk and Suraj Subramanian and Sy Choudhury and Sydney Goldman and Tal Remez and Tamar Glaser and Tamara Best and Thilo Koehler and Thomas Robinson and Tianhe Li and Tianjun Zhang and Tim Matthews and Timothy Chou and Tzook Shaked and Varun Vontimitta and Victoria Ajayi and Victoria Montanez and Vijai Mohan and Vinay Satish Kumar and Vishal Mangla and Vlad Ionescu and Vlad Poenaru and Vlad Tiberiu Mihailescu and Vladimir Ivanov and Wei Li and Wenchen Wang and Wenwen Jiang and Wes Bouaziz and Will Constable and Xiaocheng Tang and Xiaojian Wu and Xiaolan Wang and Xilun Wu and Xinbo Gao and Yaniv Kleinman and Yanjun Chen and Ye Hu and Ye Jia and Ye Qi and Yenda Li and Yilin Zhang and Ying Zhang and Yossi Adi and Youngjin Nam and Yu and Wang and Yu Zhao and Yuchen Hao and Yundi Qian and Yunlu Li and Yuzi He and Zach Rait and Zachary DeVito and Zef Rosnbrick and Zhaoduo Wen and Zhenyu Yang and Zhiwei Zhao and Zhiyu Ma},
      year={2024},
      eprint={2407.21783},
      archivePrefix={arXiv},
      primaryClass={cs.AI},
      url={https://arxiv.org/abs/2407.21783}, 
}

@inproceedings{health,
    title = "Addressing Healthcare-related Racial and {LGBTQ}+ Biases in Pretrained Language Models",
    author = "Xie, Sean  and
      Hassanpour, Saeed  and
      Vosoughi, Soroush",
    editor = "Duh, Kevin  and
      Gomez, Helena  and
      Bethard, Steven",
    booktitle = "Findings of the Association for Computational Linguistics: NAACL 2024",
    month = jun,
    year = "2024",
    address = "Mexico City, Mexico",
    publisher = "Association for Computational Linguistics",
    url = "https://aclanthology.org/2024.findings-naacl.278/",
    doi = "10.18653/v1/2024.findings-naacl.278",
    pages = "4451--4464"
}

@inproceedings{encoders,
    title = "On Measuring Social Biases in Sentence Encoders",
    author = "May, Chandler  and
      Wang, Alex  and
      Bordia, Shikha  and
      Bowman, Samuel R.  and
      Rudinger, Rachel",
    editor = "Burstein, Jill  and
      Doran, Christy  and
      Solorio, Thamar",
    booktitle = "Proceedings of the 2019 Conference of the North {A}merican Chapter of the Association for Computational Linguistics: Human Language Technologies, Volume 1 (Long and Short Papers)",
    month = jun,
    year = "2019",
    address = "Minneapolis, Minnesota",
    publisher = "Association for Computational Linguistics",
    url = "https://aclanthology.org/N19-1063/",
    doi = "10.18653/v1/N19-1063",
    pages = "622--628"
}

@article{chatgpt_bias, title={Should {C}hat{GPT} be biased? Challenges and risks of bias in large language models}, volume={28}, url={https://firstmonday.org/ojs/index.php/fm/article/view/13346}, DOI={10.5210/fm.v28i11.13346}, abstractNote={&amp;lt;p&amp;gt;As generative language models, exemplified by ChatGPT, continue to advance in their capabilities, the spotlight on biases inherent in these models intensifies. This paper delves into the distinctive challenges and risks associated with biases specifically in large-scale language models. We explore the origins of biases, stemming from factors such as training data, model specifications, algorithmic constraints, product design, and policy decisions. Our examination extends to the ethical implications arising from the unintended consequences of biased model outputs. In addition, we analyze the intricacies of mitigating biases, acknowledging the inevitable persistence of some biases, and consider the consequences of deploying these models across diverse applications, including virtual assistants, content generation, and chatbots. Finally, we provide an overview of current approaches for identifying, quantifying, and mitigating biases in language models, underscoring the need for a collaborative, multidisciplinary effort to craft AI systems that embody equity, transparency, and responsibility. This article aims to catalyze a thoughtful discourse within the AI community, prompting researchers and developers to consider the unique role of biases in the domain of generative language models and the ongoing quest for ethical AI.&amp;lt;/p&amp;gt;}, number={11}, journal={First Monday}, author={Ferrara, Emilio}, year={2023}, month={Nov.} }

@inproceedings{eval_bias,
    title = "Evaluating Interfaced {LLM} Bias",
    author = "Yeh, Kai-Ching  and
      Chi, Jou-An  and
      Lian, Da-Chen  and
      Hsieh, Shu-Kai",
    editor = "Wu, Jheng-Long  and
      Su, Ming-Hsiang",
    booktitle = "Proceedings of the 35th Conference on Computational Linguistics and Speech Processing (ROCLING 2023)",
    month = oct,
    year = "2023",
    address = "Taipei City, Taiwan",
    publisher = "The Association for Computational Linguistics and Chinese Language Processing (ACLCLP)",
    url = "https://aclanthology.org/2023.rocling-1.37/",
    pages = "292--299"
}

@misc{agent_ranking,
      title={Agent4{R}anking: Semantic Robust Ranking via Personalized Query Rewriting Using Multi-agent {LLM}}, 
      author={Xiaopeng Li and Lixin Su and Pengyue Jia and Xiangyu Zhao and Suqi Cheng and Junfeng Wang and Dawei Yin},
      year={2023},
      eprint={2312.15450},
      archivePrefix={arXiv},
      primaryClass={cs.IR},
      url={https://arxiv.org/abs/2312.15450}, 
}

@misc{deepseek,
      title={DeepSeek-{R}1: Incentivizing Reasoning Capability in {LLM}s via Reinforcement Learning}, 
      author={DeepSeek-AI and Daya Guo and Dejian Yang and Haowei Zhang and Junxiao Song and Ruoyu Zhang and Runxin Xu and Qihao Zhu and Shirong Ma and Peiyi Wang and Xiao Bi and Xiaokang Zhang and Xingkai Yu and Yu Wu and Z. F. Wu and Zhibin Gou and Zhihong Shao and Zhuoshu Li and Ziyi Gao and Aixin Liu and Bing Xue and Bingxuan Wang and Bochao Wu and Bei Feng and Chengda Lu and Chenggang Zhao and Chengqi Deng and Chenyu Zhang and Chong Ruan and Damai Dai and Deli Chen and Dongjie Ji and Erhang Li and Fangyun Lin and Fucong Dai and Fuli Luo and Guangbo Hao and Guanting Chen and Guowei Li and H. Zhang and Han Bao and Hanwei Xu and Haocheng Wang and Honghui Ding and Huajian Xin and Huazuo Gao and Hui Qu and Hui Li and Jianzhong Guo and Jiashi Li and Jiawei Wang and Jingchang Chen and Jingyang Yuan and Junjie Qiu and Junlong Li and J. L. Cai and Jiaqi Ni and Jian Liang and Jin Chen and Kai Dong and Kai Hu and Kaige Gao and Kang Guan and Kexin Huang and Kuai Yu and Lean Wang and Lecong Zhang and Liang Zhao and Litong Wang and Liyue Zhang and Lei Xu and Leyi Xia and Mingchuan Zhang and Minghua Zhang and Minghui Tang and Meng Li and Miaojun Wang and Mingming Li and Ning Tian and Panpan Huang and Peng Zhang and Qiancheng Wang and Qinyu Chen and Qiushi Du and Ruiqi Ge and Ruisong Zhang and Ruizhe Pan and Runji Wang and R. J. Chen and R. L. Jin and Ruyi Chen and Shanghao Lu and Shangyan Zhou and Shanhuang Chen and Shengfeng Ye and Shiyu Wang and Shuiping Yu and Shunfeng Zhou and Shuting Pan and S. S. Li and Shuang Zhou and Shaoqing Wu and Shengfeng Ye and Tao Yun and Tian Pei and Tianyu Sun and T. Wang and Wangding Zeng and Wanjia Zhao and Wen Liu and Wenfeng Liang and Wenjun Gao and Wenqin Yu and Wentao Zhang and W. L. Xiao and Wei An and Xiaodong Liu and Xiaohan Wang and Xiaokang Chen and Xiaotao Nie and Xin Cheng and Xin Liu and Xin Xie and Xingchao Liu and Xinyu Yang and Xinyuan Li and Xuecheng Su and Xuheng Lin and X. Q. Li and Xiangyue Jin and Xiaojin Shen and Xiaosha Chen and Xiaowen Sun and Xiaoxiang Wang and Xinnan Song and Xinyi Zhou and Xianzu Wang and Xinxia Shan and Y. K. Li and Y. Q. Wang and Y. X. Wei and Yang Zhang and Yanhong Xu and Yao Li and Yao Zhao and Yaofeng Sun and Yaohui Wang and Yi Yu and Yichao Zhang and Yifan Shi and Yiliang Xiong and Ying He and Yishi Piao and Yisong Wang and Yixuan Tan and Yiyang Ma and Yiyuan Liu and Yongqiang Guo and Yuan Ou and Yuduan Wang and Yue Gong and Yuheng Zou and Yujia He and Yunfan Xiong and Yuxiang Luo and Yuxiang You and Yuxuan Liu and Yuyang Zhou and Y. X. Zhu and Yanhong Xu and Yanping Huang and Yaohui Li and Yi Zheng and Yuchen Zhu and Yunxian Ma and Ying Tang and Yukun Zha and Yuting Yan and Z. Z. Ren and Zehui Ren and Zhangli Sha and Zhe Fu and Zhean Xu and Zhenda Xie and Zhengyan Zhang and Zhewen Hao and Zhicheng Ma and Zhigang Yan and Zhiyu Wu and Zihui Gu and Zijia Zhu and Zijun Liu and Zilin Li and Ziwei Xie and Ziyang Song and Zizheng Pan and Zhen Huang and Zhipeng Xu and Zhongyu Zhang and Zhen Zhang},
      year={2025},
      eprint={2501.12948},
      archivePrefix={arXiv},
      primaryClass={cs.CL},
      url={https://arxiv.org/abs/2501.12948}, 
}

@inproceedings{implicit_ranking,
    title = "A Study of Implicit Ranking Unfairness in Large Language Models",
    author = "Xu, Chen  and
      Wang, Wenjie  and
      Li, Yuxin  and
      Pang, Liang  and
      Xu, Jun  and
      Chua, Tat-Seng",
    editor = "Al-Onaizan, Yaser  and
      Bansal, Mohit  and
      Chen, Yun-Nung",
    booktitle = "Findings of the Association for Computational Linguistics: EMNLP 2024",
    month = nov,
    year = "2024",
    address = "Miami, Florida, USA",
    publisher = "Association for Computational Linguistics",
    url = "https://aclanthology.org/2024.findings-emnlp.467/",
    doi = "10.18653/v1/2024.findings-emnlp.467",
    pages = "7957--7970"
}

@inproceedings{better_agents,
    author = {Sun, Guangzhi and Zhan, Xiao and Such, Jose},
    title = {Building Better {AI} Agents: A Provocation on the Utilisation of Persona in {LLM}-based Conversational Agents},
    year = {2024},
    isbn = {9798400705113},
    publisher = {Association for Computing Machinery},
    address = {New York, NY, USA},
    url = {https://doi.org/10.1145/3640794.3665887},
    doi = {10.1145/3640794.3665887},
    booktitle = {Proceedings of the 6th ACM Conference on Conversational User Interfaces},
    articleno = {35},
    numpages = {6},
    location = {Luxembourg, Luxembourg},
    series = {CUI '24}
    }

@inproceedings{creators,
author = {Choi, Yoonseo and Kang, Eun Jeong and Choi, Seulgi and Lee, Min Kyung and Kim, Juho},
title = {Proxona: Supporting Creators' Sensemaking and Ideation with {LLM}-Powered Audience Personas},
year = {2025},
isbn = {9798400713941},
publisher = {Association for Computing Machinery},
address = {New York, NY, USA},
url = {https://doi.org/10.1145/3706598.3714034},
doi = {10.1145/3706598.3714034},
booktitle = {Proceedings of the 2025 CHI Conference on Human Factors in Computing Systems},
articleno = {149},
numpages = {32},
location = {
},
series = {CHI '25}
}

@inproceedings{temp,
    title = "The Effect of Sampling Temperature on Problem Solving in Large Language Models",
    author = "Renze, Matthew",
    editor = "Al-Onaizan, Yaser  and
      Bansal, Mohit  and
      Chen, Yun-Nung",
    booktitle = "Findings of the Association for Computational Linguistics: EMNLP 2024",
    month = nov,
    year = "2024",
    address = "Miami, Florida, USA",
    publisher = "Association for Computational Linguistics",
    url = "https://aclanthology.org/2024.findings-emnlp.432/",
    doi = "10.18653/v1/2024.findings-emnlp.432",
    pages = "7346--7356"
}

@article{deepmath,
  title={DeepMath-103K: A Large-Scale, Challenging, Decontaminated, and  Verifiable Mathematical Dataset for Advancing Reasoning},
  author={He, Zhiwei and Liang, Tian and Xu, Jiahao and Liu, Qiuzhi and Chen, Xingyu and Wang, Yue and Song, Linfeng and Yu, Dian and Liang, Zhenwen and Wang, Wenxuan and Zhang, Zhuosheng and Wang, Rui and Tu, Zhaopeng and Mi, Haitao and Yu, Dong},
  year={2025},
  eprint={2504.11456},
  archivePrefix={arXiv},
  primaryClass={cs.CL},
  url={https://arxiv.org/abs/2504.11456}, 
}

@inproceedings{mathqa,
    title = "{M}ath{QA}: Towards Interpretable Math Word Problem Solving with Operation-Based Formalisms",
    author = "Amini, Aida  and
      Gabriel, Saadia  and
      Lin, Shanchuan  and
      Koncel-Kedziorski, Rik  and
      Choi, Yejin  and
      Hajishirzi, Hannaneh",
    editor = "Burstein, Jill  and
      Doran, Christy  and
      Solorio, Thamar",
    booktitle = "Proceedings of the 2019 Conference of the North {A}merican Chapter of the Association for Computational Linguistics: Human Language Technologies, Volume 1 (Long and Short Papers)",
    month = jun,
    year = "2019",
    address = "Minneapolis, Minnesota",
    publisher = "Association for Computational Linguistics",
    url = "https://aclanthology.org/N19-1245/",
    doi = "10.18653/v1/N19-1245",
    pages = "2357--2367"
}

@inproceedings{
  llm-mcq-bias,
  title={Large Language Models Are Not Robust Multiple Choice Selectors},
  author={Chujie Zheng and Hao Zhou and Fandong Meng and Jie Zhou and Minlie Huang},
  booktitle={The Twelfth International Conference on Learning Representations},
  year={2024},
  url={https://openreview.net/forum?id=shr9PXz7T0}
}

@misc{investigating,
      title={Investigating Implicit Bias in Large Language Models: A Large-Scale Study of Over 50 LLMs}, 
      author={Divyanshu Kumar and Umang Jain and Sahil Agarwal and Prashanth Harshangi},
      year={2024},
      eprint={2410.12864},
      archivePrefix={arXiv},
      primaryClass={cs.CL},
      url={https://arxiv.org/abs/2410.12864}, 
}

\appendix

\section{Appendix}
\label{sec:appendix}

\begin{table}[h]
  \centering
  \begin{tabular}{lll}
    \hline
    System & \vline & \begin{minipage}[t]{0.65\columnwidth}
    You are a helpful American assistant. Make sure that your responses closely embody your identity. \end{minipage} \\
    \hline
    User & \vline & \begin{minipage}[t]{0.65\columnwidth} 
    Please answer the following multiple-choice question carefully. Only answer with a single letter. Do not respond with any other text, numbers, or symbols.

    <QUESTION>
    \end{minipage}\\
    \hline
    Assistant & \vline & \begin{minipage}[t]{0.65\columnwidth} 
    <RESPONSE>\end{minipage}\\
    \hline
  \end{tabular}
  \caption{Example conversation with chat role and baseline prompt used in the experiment for GSM-MC and MathQA.}
  \label{tab:example1}
\end{table}

\begin{table}[h]
  \centering
  \begin{tabular}{lll}
    \hline
    System & \vline & \begin{minipage}[t]{0.65\columnwidth}
    You are a helpful American assistant. Make sure that your responses closely embody your identity. \end{minipage} \\
    \hline
    User & \vline & \begin{minipage}[t]{0.65\columnwidth} 
    Please answer the following question carefully. Only respond with the final answer. Do not respond with any other text. Output the final answer within \textbackslash boxed\{\}.

    <QUESTION>
    \end{minipage}\\
    \hline
    Assistant & \vline & \begin{minipage}[t]{0.65\columnwidth} 
    <RESPONSE>\end{minipage}\\
    \hline
  \end{tabular}
  \caption{Example conversation with chat role and baseline prompt used in the experiment for DeepMath.}
  \label{tab:example2}
\end{table}

\begin{table*}
  \centering
  \begin{tabular}{l|cccccc}
    \hline
    % & Llama3 & Llama3.1 & Llama3.1& Llama3.2 & Llama3.3 & Mistral & Phi & Gemma \\
    % \hline
    % Parameters & 8B & 8B & 70B & 3B & 70B & 7B & 3.8B & 9B \\
    \textbf{Models} & Llama 3.1 & Llama 3.2 & Llama 3.3 & Mistral v0.3 & Phi 3.5 & Gemma 2 \\
    % \hline
    \hline
    \textit{Model Parameters} & 8B & 3B & 70B & 7B & 3.8B & 9B \\
    % \hline
    \hline
    $t=0.2$ & 71.7 & 69.1 & 94.4 & 48.7 & 78.9 & 90.7 \\ 
    $t=0.4$ & 50.8 & 17.8 & 93.8 & 30.3 & 72.6 & 87.9 \\ 
    $t=0.6$ & 25.7 & 0 & 93.0 & 10.8 & 56.1 & 86.0 \\
    $t=0.8$ & 0 & 0 & 92.1 & 0 & 39.1 & 84.1 \\
    $t=1.0$ & 0 & 0 & 90.6 & 0 & 16.8 & 79.5 \\
    \hline
  \end{tabular}
  \caption{DIF (GSM-MC) scores of different models across different temperatures.}
  \label{tab:vanilla_results_temp}
\end{table*}

\subsection{Temperature and bias}
\label{sec:temp}

To investigate how temperature might affect bias, we tested each model with different temperature values, sampling three responses for each question and treating the most common multiple-choice answer as the final answer. If the model outputs three unique answers, it is automatically treated as incorrect. As expected in Table~\ref{tab:vanilla_results_temp}, allowing random sampling introduces a substantial amount of variance in answers, resulting in higher bias with higher temperatures. Given the observation that implicit bias and intelligence are inversely correlated, and previous research that observes a lack of significant influence of temperature on problem solving, it follows that temperature might not have much of an impact on implicit bias \cite{temp}, and the increasing bias is solely the result of random sampling.

\begin{table*}[h]
  \centering
  \begin{tabular}{l|cccccc}
    \hline
    \textbf{Models} & Llama 3.1 & Llama 3.2 & Llama 3.3 & Mistral v0.3 & Phi 3.5 & Gemma 2 \\
    % \hline
    \hline
    \textit{Model Parameters} & 8B & 3B & 70B & 7B & 3.8B & 9B \\
    \hline
    \hline
    Baseline Persona & 356 & 159 & 597 & 258 & 362 & 472 \\
    % \hline
    \hline
    American Indian & 372 & 161 & 593 & 232 & 356 & 474 \\
    Asian & 374 & 164 & 592 & 224 & 362 & 463 \\
    Black & 369 & 167 & 598 & 224 & 359 & 475 \\
    Hispanic & 370 & 163 & 598 & 209 & 356 & 467 \\
    Middle Eastern & 361 & 155 & 594 & 221 & 360 & 468\\
    Pacific Islander & 366 & 162 & 591 & 207 & 359 & 473 \\
    White & 361 & 155 & 590 & 243 & 360 & 474 \\
    % \hline
    \hline
    Atheist & 360 & 160 & 590 & 244 & 357 & 470 \\
    Buddhist & 360 & 152 & 597 & 218 & 355 & 477 \\
    Christian & 366 & 169 & 596 & 243 & 358 & 472 \\
    Hindu & 361 & 159 & 593 & 234 & 352 & 477 \\
    Jewish & 368 & 165 & 591 & 219 & 359 & 470 \\
    Mormon & 365 & 164 & 591 & 243 & 365 & 474 \\
    Muslim & 366 & 158 & 598 & 215 & 360 & 472 \\
    % \hline
    \hline
    Female & 355 & 164 & 593 & 252 & 359 & 474 \\
    Male & 353 & 167 & 599 & 267 & 367 & 476 \\
    Non-binary & 364 & 165 & 599 & 231 & 356 & 466 \\
    % \hline
    \hline
    Gay & 371 & 162 & 584 & 248 & 360 & 467 \\
    Straight & 351 & 160 & 591 & 258 & 362 & 469 \\
    % \hline
    \hline
    Able-bodied & 354 & 158 & 588 & 249 & 361 & 465 \\
    Physically disabled & 366 & 165 & 597 & 218 & 358 & 469 \\
    \hline
  \end{tabular}
  \caption{Correct answers out of 1000 for the vanilla testing of personas when greedy decoding is used for text generation on GSM-MC.}
  \label{tab:vanilla_results_top_k}
\end{table*}

\begin{table*}[h]
  \centering
  \begin{tabular}{l|cccccc}
    \hline
    \textbf{Models} & Llama 3.1 & Llama 3.2 & Llama 3.3 & Mistral v0.3 & Phi 3.5 & Gemma 2 \\
    % \hline
    \hline
    \textit{Model Parameters} & 8B & 3B & 70B & 7B & 3.8B & 9B \\
    \hline
    \hline
    Baseline Persona & 182 & 177 & 195 & 198 & 99 & 159  \\
    % \hline
    \hline
    American Indian & 188 & 167 & 216 & 196 & 116 & 161   \\
    Asian & 187 & 191 & 211 & 193 & 107 & 164  \\
    Black & 187 & 191 & 206 & 191 & 107 & 167 \\
    Hispanic & 185 & 184 & 211 & 186 & 112 & 163 \\
    Middle Eastern & 186 & 186 & 202 & 189 & 114 & 155\\
    Pacific Islander & 183 & 193 & 206 & 193 & 113 & 162  \\
    White & 169 & 190 & 201 & 188 & 111 & 155  \\
    % \hline
    \hline
    Atheist & 169 & 161 & 207 & 190 & 110 & 160  \\
    Buddhist & 181 & 170 & 205 & 191 & 112 & 152  \\
    Christian & 191 & 191 & 209 & 199 & 109 & 169 \\
    Hindu & 186 & 170 & 204 & 191 & 117 & 159 \\
    Jewish & 180 & 175 & 210 & 182 & 112 & 165 \\
    Mormon & 192 & 182 & 205 & 188 & 113 & 164  \\
    Muslim & 189 & 177 & 202 & 188 & 112 & 158 \\
    % \hline
    \hline
    Female & 189 & 192 & 202 & 193 & 110 & 164 \\
    Male & 183 & 186 & 207 & 193 & 100 & 167  \\
    Non-binary & 171 & 133 & 214 & 193 & 110 & 165  \\
    % \hline
    \hline
    Gay & 176 & 188 & 204 & 181 & 107 & 162  \\
    Straight & 180 & 192 & 201 & 189 & 104 & 160  \\
    % \hline
    \hline
    Able-bodied & 169 & 178 & 201 & 186 & 113 & 158  \\
    Physically disabled & 178 & 192 & 186 & 165 & 111 & 168  \\
    \hline
  \end{tabular}
  \caption{Correct answers out of 1000 for the vanilla testing of personas when greedy decoding is used for text generation for MathQA.}
  \label{tab:vanilla_results_top_k}
\end{table*}

\begin{table*}[h]
  \centering
  \begin{tabular}{l|cccccc}
    \hline
    \textbf{Models} & Llama 3.1 & Llama 3.2 & Llama 3.3 & Mistral v0.3 & Phi 3.5 & Gemma 2 \\
    % \hline
    \hline
    \textit{Model Parameters} & 8B & 3B & 70B & 7B & 3.8B & 9B \\
    \hline
    \hline
    Baseline Persona & 49 & 39 & 55 & 33 & 29 & 34  \\
    % \hline
    \hline
    American Indian & 47 & 42 & 54 & 30 & 27 & 35   \\
    Asian & 47 & 37 & 52 & 31 & 35 & 33  \\
    Black & 49 & 41 & 53 & 31 & 34 & 35 \\
    Hispanic & 48 & 38 & 52 & 31 & 36 & 35 \\
    Middle Eastern & 48 & 38 & 52 & 32 & 34 & 31 \\
    Pacific Islander & 49 & 40 & 55 & 27 & 36 & 35  \\
    White & 49 & 42 & 57 & 33 & 36 & 35  \\
    % \hline
    \hline
    Atheist & 49 & 42 & 54 & 36 & 40 & 36  \\
    Buddhist & 49 & 43 & 55 & 30 & 39 & 37  \\
    Christian & 52 & 43 & 55 & 32 & 30 & 35 \\
    Hindu & 49 & 38 & 52 & 29 & 35 & 36 \\
    Jewish & 48 & 41 & 55 & 33 & 33 & 34 \\
    Mormon & 51 & 40 & 55 & 33 & 30 & 36  \\
    Muslim & 50 & 42 & 54 & 31 & 36 & 34 \\
    % \hline
    \hline
    Female & 47 & 38 & 55 & 33 & 28 & 40 \\
    Male & 50 & 37 & 56 & 32 & 35 & 35  \\
    Non-binary & 46 & 40 & 55 & 33 & 38 & 35  \\
    % \hline
    \hline
    Gay & 47 & 44 & 55 & 30 & 34 & 36  \\
    Straight & 49 & 41 & 56 & 34 & 32 & 34  \\
    % \hline
    \hline
    Able-bodied & 48 & 42 & 57 & 32 & 35 & 31  \\
    Physically disabled & 48 & 42 & 54 & 33 & 38 & 34  \\
    \hline
  \end{tabular}
  \caption{Correct answers out of 1000 for the vanilla testing of personas when greedy decoding is used for text generation for DeepMath.}
  \label{tab:vanilla_results_top_k}
\end{table*}

\end{document}